\documentclass[5p,times,twocolumn]{elsarticle}

\usepackage{amsmath,amssymb,amsfonts,color}
\usepackage{graphicx}
\usepackage{textcomp}
\usepackage{xcolor}
\usepackage{enumerate}
\usepackage{setspace}
\usepackage{bm}
\usepackage{comment}
\usepackage{caption}
\usepackage{subcaption}
\usepackage{tipa}
\usepackage{footnote}
\usepackage{algorithm}
\usepackage{algorithmic}
\usepackage{url}
\usepackage{multirow}

\def\x{{\mathbf x}}

\usepackage[dvipsnames]{xcolor}
\begin{document}

\begin{frontmatter}

\title{Dual-Model Weight Selection and Self-Knowledge Distillation for \\ Medical Image Classification}

\author{Ayaka Tsutsumi${}^\text{a}$}
\ead{tsutsumi@lmd.ist.hokudai.ac.jp}
\author{Guang Li${}^\text{a}$}
\ead{guang@lmd.ist.hokudai.ac.jp}
\author{Ren Togo${}^\text{a}$}
\ead{togo@lmd.ist.hokudai.ac.jp}
\author{Takahiro Ogawa${}^\text{a}$}
\ead{ogawa@lmd.ist.hokudai.ac.jp}
\author{Satoshi Kondo${}^\text{b}$}
\ead{kondo@muroran-it.ac.jp}
\author{Miki Haseyama${}^\text{a}$}
\ead{mhaseyama@lmd.ist.hokudai.ac.jp}
\address{${}^\text{a}$Hokkaido University, \\
           N-14, W-9, Kita-Ku, Sapporo, Hokkaido, 060-0814, Japan}
\address{${}^\text{b}$Muroran Institute of Technology, \\
           27-1, Mizumoto-cho, Muroran, Hokkaido, 050-8585, Japan}

\begin{abstract}

We propose a novel medical image classification method that integrates dual-model weight selection with self-knowledge distillation (SKD).
In real-world medical settings, deploying large-scale models is often limited by computational resource constraints, which pose significant challenges for their practical implementation. Thus, developing lightweight models that achieve comparable performance to large-scale models while maintaining computational efficiency is crucial.
To address this, we employ a dual-model weight selection strategy that initializes two lightweight models with weights derived from a large pretrained model, enabling effective knowledge transfer.
Next, SKD is applied to these selected models, allowing the use of a broad range of initial weight configurations without imposing additional excessive computational cost, followed by fine-tuning for the target classification tasks.
By combining dual-model weight selection with self-knowledge distillation, our method overcomes the limitations of conventional approaches, which often fail to retain critical information in compact models.
Extensive experiments on publicly available datasets---chest X-ray images, lung computed tomography scans, and brain magnetic resonance imaging scans---demonstrate the superior performance and robustness of our approach compared to existing methods.
\end{abstract}


\end{frontmatter}
\section{Introduction}
\label{sec1}
Deep learning has become indispensable in the medical field, driving significant advances in diagnostic imaging, disease prediction, and electronic health record analysis~\cite{wang2019deep, tasai_icassp_2025}. Large-scale models have achieved remarkable success across diverse tasks; however, their deployment in clinical settings is often hindered by limited computational resources~\cite{li2022self, aouedi2022handling}. For example, analyzing high-resolution medical images such as chest X-rays and magnetic resonance imaging (MRI) scans requires substantial computing power and memory, making large model deployment impractical in many healthcare environments~\cite{li2022covid, li2023boosting, li2024large}. Simultaneously, the growing shortage of healthcare professionals is increasing the need for efficient automated systems to reduce workloads~\cite{li2023self, corral2024energy}. Consequently, there is a rising demand for compact models that can operate effectively under resource constraints while maintaining performance comparable to large-scale counterparts. \par
A critical factor in improving small-model performance is weight initialization. Poor initialization can lead to vanishing or exploding gradients, hindering effective training. While commonly used methods such as Xavier~\cite{pmlr-v9-glorot10a} and Kaiming~\cite{he2015delving} initialization support stable training from scratch, pretrained large-scale models (e.g., ImageNet-21K~\cite{deng2009imagenet} or LAION-5B~\cite{schuhmann2022laion}) are increasingly adopted for transfer learning. However, fine-tuning these large models is often infeasible in clinical settings because it demands high-end GPUs with large memory and substantial computing resources---conditions rarely met in real-world medical environments. To address this, recent studies have explored weight selection methods, which transfer only a subset of weights from a large pretrained model to a smaller one~\cite{xu2024initializing, xia2024exploring, tsutsumi2024lung}. \par
Weight selection aims to enable efficient knowledge transfer by extracting and using a subset of weights from a large pretrained model to initialize a smaller model~\cite{xu2024initializing}. Compared with full fine-tuning, this approach reduces computational cost~\cite{grigore2024weight, han2016deep, frankle2019lottery}. However, existing methods face two main challenges: the limited capacity of small models restricts the amount of transferable knowledge~\cite{chen2020survey, zhang2017understanding}, and a one-time selection process may overlook important weights, limiting performance gains. \par
A straightforward idea to address these challenges is to use two small models initialized with complementary subsets of weights from the large model. This approach can capture a broad range of information and offset individual model limitations. However, training two models simultaneously increases resource demands, conflicting with the goal of lightweight deployment. Therefore, it is desirable to design a strategy that retains the advantages of dual initialization without increasing graphics processing unit (GPU) memory usage~\cite{lan2020albert}. \par
In this study, we propose a novel medical image classification approach based on dual-model weight selection and self-knowledge distillation (SKD). Our method initializes two small models with identical architectures using different subsets of weights from a large pretrained model (dual-model weight selection). One model serves as the trainable main model, and the other as a fixed auxiliary model. SKD is then performed, where the main model learns from the ground truth labels and soft targets generated by the auxiliary model. This allows the main model to benefit from complementary information from both weight subsets, improving learning performance without additional excessive memory consumption. \par
Extensive experiments on chest X-ray images, lung computed tomography (CT) scans, and brain MRI datasets show that our method significantly boosts the performance of compact models. By enriching the initialization process with dual-model weight selection and leveraging SKD, the main model gains complementary insights without extra memory cost. This enhances representation learning and mitigates small-model limitations, resulting in higher accuracy while preserving efficiency. Our approach offers a practical and scalable solution for real-world, resource-constrained medical settings. \par
Our key contributions are as follows:
\begin{itemize}
    \item We propose a dual-model weight selection method in which two small models are initialized with different subsets of weights from a large pretrained model, enabling broader knowledge transfer without imposing excessive computational cost.
    \item We introduce a SKD mechanism where a fixed auxiliary model provides soft targets to guide the main model, effectively mitigating the capacity limitations of small models.
    \item We validate the proposed method through extensive experiments on multiple publicly available datasets across diverse imaging modalities---chest X-rays, lung CT scans, and brain MRIs---demonstrating consistent outperformance over baseline approaches.
\end{itemize}
The remainder of this paper is structured as follows. Section 2 reviews related work. Section 3 details the proposed method. Sections 4 and 5 present experimental results and discussion, respectively. Finally, Section 6 concludes the paper.

\section{Related Work}
\label{sec2}
\subsection{Parameter Initialization and Weight Selection}
\label{2.1}
Parameter initialization is crucial for both the convergence speed and generalization performance of deep learning models. Common techniques, such as Xavier initialization~\cite{pmlr-v9-glorot10a} and Kaiming initialization~\cite{he2015delving}, scale weights based on the input and output sizes of layers to prevent gradient vanishing and exploding gradients. While conventional methods rely on random initialization, the rise of pretrained models has introduced alternative strategies that leverage pretrained parameter distributions. For example, Lin et al.~\cite{lin2021weight} adapted weights of large models to smaller ones using trainable transformations, while Sanh et al.~\cite{sanh2019distilbert} demonstrated effective initialization by selecting layers from pre-trained BERT models. \par
Recent efforts also explore pretrained weight reuse based on layer-wise alignment or similarity-based matching techniques~\cite{chen2023efficient}, as well as universal initialization schemes tailored for Vision Transformers~\cite{han2021transformer}. Other methods examine the statistical properties of pretrained models, such as initializing convolutional layers using covariance structures~\cite{trockman2023covariance} or transferring self-attention patterns~\cite{trockman2023mimetic}. These trends show that initialization has moved from random heuristics to data-driven and architecture-aware strategies~\cite{neyshabur2020what, kolesnikov2020bit}. \par
In contrast, the recently proposed weight selection methods~\cite{xu2024initializing, xia2024exploring} represent a novel direction for initializing smaller models by selectively extracting parameters from large pre-trained models. This method enables efficient knowledge transfer by utilizing selected weights, leading to faster convergence and reduced training costs relative to models initialized from scratch~\cite{grigore2024weight}. Saliency-guided selection~\cite{wang2023saliency} and structured sparsity-aware techniques~\cite{ainslie2020etc} further enhance the quality of the selected weights, while task-specific fine-tuning strategies have also been introduced to mitigate performance degradation in low-resource scenarios~\cite{kumar2022fine}.
\subsection{Knowledge Distillation}
\label{2.2}
Knowledge distillation is a technique for compressing and optimizing deep learning models by transferring knowledge from a complex teacher model to a smaller, efficient student model~\cite{gou2021knowledge}. Following the work of Hinton et al.~\cite{hinton2015distilling}, this technique gained widespread attention, leading to the development of various knowledge distillation approaches that focus on output responses, feature maps, and interfeature relationships. In response-based distillation, the student model learns to mimic the output probabilities (soft targets) of the teacher model, capturing interclass relationships and enhancing generalization performance~\cite{song2023exploring, yuan2020revisiting}. Feature-based distillation transfers intermediate layer feature maps~\cite{ji2021show}, as demonstrated in FitNets~\cite{romero2015fitnets}, while relation-based distillation leverages structural relationships among data points~\cite{dong2021few, tian2020contrastive}.
To bridge capacity gaps between large and small models, methods such as teacher assistant distillation~\cite{mirzadeh2020improved} and online collaborative learning~\cite{wang2021online} have been proposed. Distillation under a distribution shift remains an emerging challenge~\cite{allen2023kdshift}. \par
SKD eliminates the need for an external teacher by enabling the model to improve itself. Zhang et al.~\cite{zhang2020class} introduced a method for segmenting deep models such as ResNet into blocks, allowing intra-model knowledge transfer between components. Kim et al.~\cite{kim2021self} introduced progressive SKD, and Ge et al.~\cite{ge2021self} developed batch-wise ensembling of self-predictions to suppress incorrect predictions. Recent work has adapted these ideas for transformers~\cite{yuan2023self} and even explored self-distillation in the absence of ground-truth labels using self-supervised techniques~\cite{zhang2021label}.
MiniLM~\cite{wang2020minilm} exemplifies a task-agnostic SKD method that maintains performance while significantly reducing the model size.
Beyond image classification, KD and SKD have also been extended to more complex visual tasks. In 3D vision, Cao et al.~\cite{cao2023kt} formulated unpaired shape completion as a knowledge-transfer problem using a teacher–assistant–student framework. In remote sensing, Du et al.~\cite{du2024lumnet} injected land-use knowledge into a multiscale CNN for single-image height estimation, while Li et al.~\cite{li2023patch} proposed a patch-similarity-based self-distillation scheme that improves cross-view geo-localization by distilling multi-scale local cues. These applications demonstrate that distillation can effectively enhance representation learning under domain shifts and structural ambiguities, which is closely related to our goal of robust medical image classification.
A comprehensive survey on knowledge distillation summarizes these trends and outlines the future directions~\cite{beyer2022knowledge}.
\begin{figure}[t]
        \centering
        \includegraphics[width=8cm]{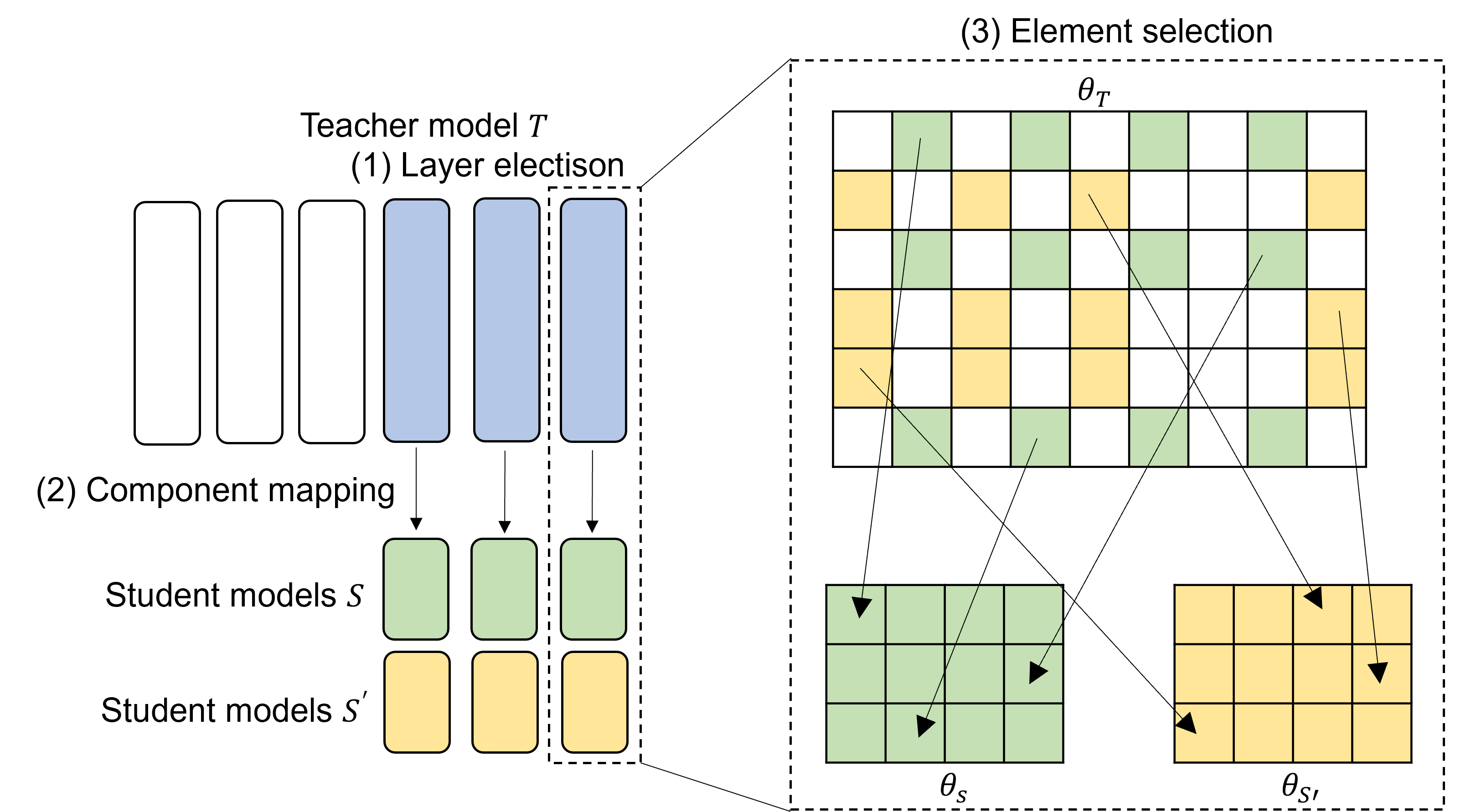}
        \caption{Overview of the proposed dual-model weight selection method. The teacher model $T$ is used to initialize the main student model $S$ and the auxiliary student model $S'$.}
        \label{fig1}
\end{figure}
\section{Dual-model Weight Selection and Self-knowledge Distillation}
\label{sec3}
Our method employs a pretrained large model as the teacher and small-scale models as students, integrating dual-model weight selection for effective knowledge transfer and SKD for efficient training. First, weight selection transfers essential parameters from the teacher model to the student models, enabling them to inherit prior knowledge and achieve strong early performance. Then, SKD allows the student models to refine their learning by leveraging their own predictions, improving task performance, and maximizing the benefit of the initial weights without imposing a high computational cost.
Traditional pruning or saliency-driven subset selection aims primarily at parameter reduction, often at the cost of representational diversity. In contrast, our method derives two complementary parameter subsets that strictly preserve structural index alignment, ensuring that both networks retain coherent functional mappings while encouraging diversified feature specialization. Moreover, rather than relying on a separate and typically larger teacher model as in standard teacher–student distillation, we adopt an EMA-updated auxiliary student as a self-adaptive teacher. This design provides stable supervision signals and prevents training collapse, while avoiding additional backpropagation overhead, thus achieving both efficiency and performance gains.
\subsection{Dual-model Weight Selection}
\label{3.1}

We first perform dual-model weight selection. As shown in Fig.~\ref{fig1}, weights are selected from the teacher model to initialize two student models for subsequent SKD. Here, the teacher model is denoted as $T$, the main student model as $S$, and the auxiliary student model as $S'$. The main and auxiliary models share the same architecture and can be interchanged. The weight selection process involves three steps: (1) layer selection, (2) component mapping, and (3) element selection. \par
\textbf{(1) Layer selection.} In this step, each layer of the student model is initialized with weights from the corresponding layer of the teacher model. For isotropic network architectures such as ViT~\cite{dosovitskiy2021image} and MLP-Mixer~\cite{tolstikhin2021mlp}, the first $N$ layers of the teacher model are selected, where $N$ is the total number of layers in the student model. For hierarchical architectures like Swin Transformer~\cite{liu2021swin} and ConvNeXt~\cite{liu2022convnet}, where scales and embedding dimensions vary across layers, or for classic convolutional networks like VGG~\cite{simonyan2014two}, which capture multi-scale features by progressively reducing spatial dimensions and increasing channel dimensions, layer selection involves taking the first-$N$ layers from each hierarchical stage.\par
\textbf{(2) Component mapping.} In this step, the weights of the selected layers are transferred from the teacher model to the corresponding layers of the student models. Modern neural networks are typically modular, with layers in models of the same family sharing identical component structures that differ mainly in width. This design enables straightforward one-to-one mapping. After mapping, the components of student models $S$ and $S'$ are initialized with weights from the teacher model $T$. Despite this common origin, $S$ and $S'$ maintain different weight parameters, enabling them to learn independently. \par
\begin{figure}[t]
        \centering
        \includegraphics[width=8cm]{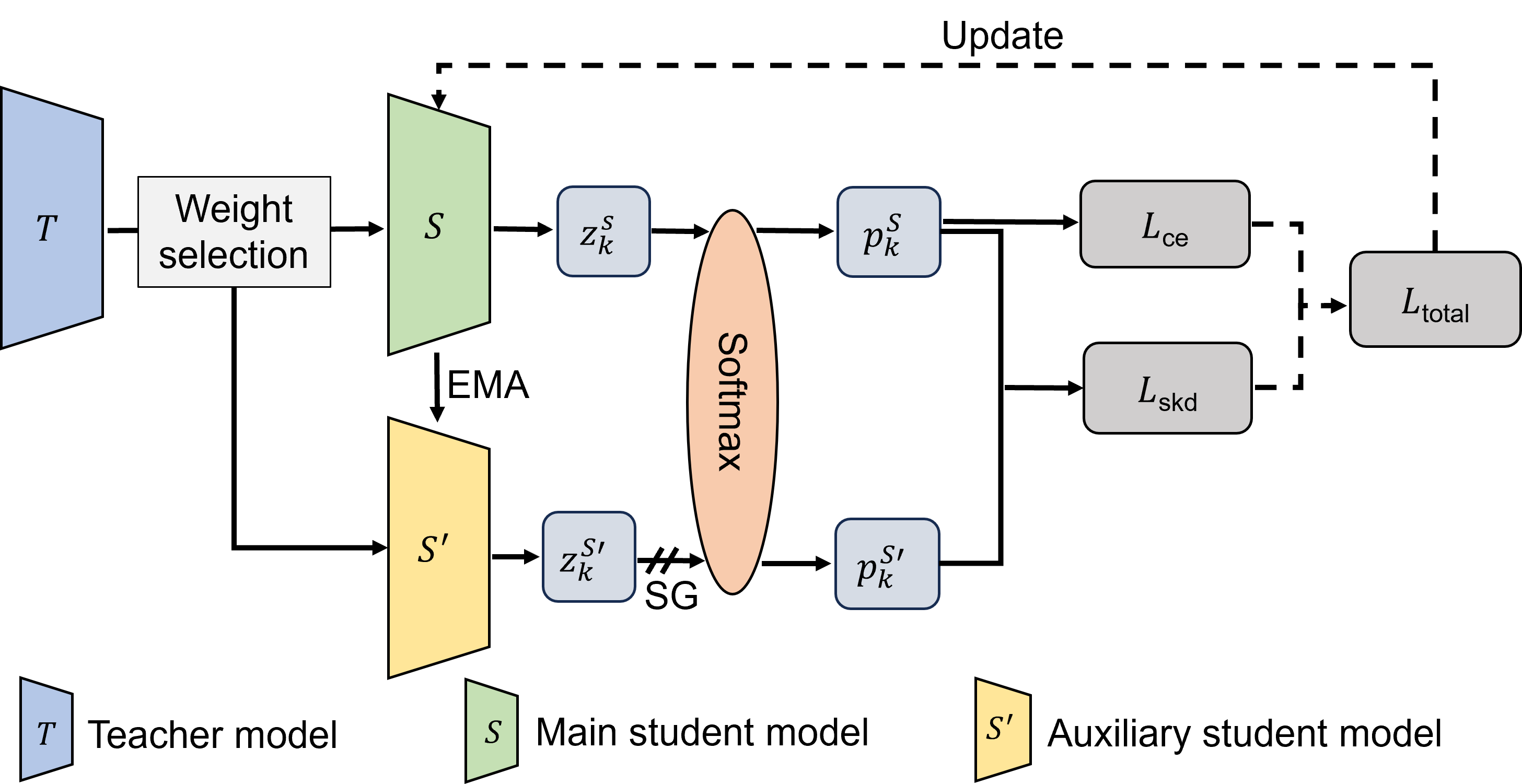}
        \caption{Overview of the proposed self-knowledge distillation method. The weights of the auxiliary student model $S'$ are the exponential moving average (EMA) of the weights of the main student model $S$, and SG denotes the stop gradient operation.}
        \label{fig2}
\end{figure}
\begin{table*}[t]
    \centering
    \caption{Architecture details of the teacher and student models used in this study.}
    \setlength{\tabcolsep}{18pt}
    \begin{tabular}{l|cc|cc}
        \hline
        Configuration & \multicolumn{2}{c|}{Teacher model} & \multicolumn{2}{c}{Student model} \\
        \hline
        Model & ViT-S~\cite{dosovitskiy2021image} & ConvNeXt-T~\cite{liu2022convnet} & ViT-T~\cite{dosovitskiy2021image} & ConvNeXt-F~\cite{liu2022convnet} \\
        Depth & 12 & 3 / 3 / 9 / 3 & 12 & 2 / 2 / 6 / 2 \\
        Embedding dimension & 384 & 96 / 192 / 384 / 768 & 192 & 48 / 96 / 192 / 384 \\
        Number of heads & 6 & - & 3 & - \\
        Number of parameters & 22M & 28M & 5M & 5M \\
        \hline
    \end{tabular}
    \label{tab1}
\end{table*}
\textbf{(3) Element selection.} When initializing the weights $\theta_S$ and $\theta_{S'}$ in the student models using the teacher model's weights $\theta_T$, consistent indices are selected across all weight tensors within each student model to preserve internal structure. This consistency ensures that the relationships and interactions captured in the teacher model are maintained in both student models, enabling effective knowledge transfer~\cite{xu2024initializing}. Furthermore, by selecting different subsets of elements for each student model---while maintaining internal index consistency---the models focus on different regions of the teacher's
parameter space. This strategy promotes diverse knowledge representation among students while preserving structural integrity within each model, leading to more comprehensive knowledge transfer. 
By distributing different elements among the student models, the method encourages specialization and reduces redundancy, ultimately enhancing overall performance. Thus, distinct but internally consistent subsets of weights are selected from $\theta_T$ to initialize $\theta_S$ and $\theta_{S'}$, allowing each student model to inherit and specialize in different aspects of the teacher model's knowledge while collectively capturing a broad and diverse representation.\par
The main student model $S$ can then be fine-tuned on specific medical image datasets, leveraging the support of the auxiliary student model $S'$. Through fine-tuning, the main student model $S$ learns dataset-specific representations for downstream tasks such as classification. By initializing the student models $S$ and $S'$ with selected weights from the teacher model $T$, we improve the accuracy of medical image classification while maintaining a lightweight architecture.
\subsection{Self-knowledge Distillation}
\label{3.2}
An overview of the SKD process is presented in Fig.~\ref{fig2}. Following weight selection, we obtain the main and auxiliary student models $S$ and $S'$, respectively. Through SKD, knowledge from $S$ and $S'$ is utilized within a shared network structure, allowing efficient information transfer without incurring additional computational resources. First, the predicted class probabilities from the main student model $S$ are computed by applying the softmax function $M$ to its output logits as follows:
\begin{equation}
p_k^{S} = \frac{\exp\left(\frac{z_k^{S}}{\tau}\right)}{\sum_{k=1}^{K} \exp\left(\frac{z_k^{S}}{\tau}\right)}
,
\end{equation}
where $z_k^{S}$ represents the $k$-th output logits of $S$, $K$ is the number of classes, and $\tau$ is a temperature parameter.
Similarly, the predicted probabilities of the auxiliary student model $S'$ are obtained as follows:
\begin{equation}
p_k^{S'} = \frac{\exp\left(\frac{z_k^{S'}}{\tau}\right)}{\sum_{k=1}^{K} \exp\left(\frac{z_k^{S'}}{\tau}\right)}
.
\end{equation}
Using these predicted probabilities, the SKD loss $L_{\text{skd}}$ is calculated as follows:
\begin{equation}
L_{\text{skd}} = \sum_{k=1}^{K} p_k^{S'} \log\left(\frac{p_k^{S'}}{p_k^{S}}\right)
.\end{equation}
Here, $p_k^{S'}$ represents the soft targets from $S'$, and $p_k^{S}$ represents the output of $S$. 
The SKD loss is calculated by comparing the outputs of the two student models, allowing $S$ to learn from the auxiliary student models $S'$. Since the gradient is not backpropagated from $S'$, this prevents additional training overhead, allowing the model to improve accuracy with minimal computational cost. Additionally, the cross-entropy loss $L_{\text{ce}}$ for the output logits of $S$ is calculated as follows:
\begin{equation}
L_{\text{ce}} = -\sum_{k=1}^{K} y_k \log(p_k^{S})
,\end{equation}
where $y_k$ is the one-hot encoded target label, and $p_k^{S}$ is the softmax output of the main student model $S$. The cross-entropy loss measures the discrepancy between the predicted class probabilities and the ground truth labels, serving as the primary objective for the classification task.
Finally, the total loss $L_{\text{total}}$ is calculated by combining the SKD loss $L_{\text{skd}}$ and the cross-entropy loss $L_{\text{ce}}$ as follows:
\begin{equation}
L_{\text{total}} = \alpha L_{\text{ce}} + (1 - \alpha) \tau^2 L_{\text{skd}}
.\end{equation}
Here, $\alpha$ is a hyperparameter that controls the trade-off between the cross-entropy loss and the distillation loss. The total loss, denoted as $L_{\text{total}}$, allows the model to learn from both the ground truth labels and the distilled knowledge. The weights of model \( S \) are updated based on the calculated total loss \( L_{\text{total}} \) as follows:
\begin{equation}
\theta_S \leftarrow \theta_S - \eta \nabla_{\theta_S} L_{\text{total}},
\end{equation}
where \( \eta \) denotes the learning rate, and \( \theta_S \) represents the parameters of the main student model \( S \). Then, the weights of the auxiliary student model \( S' \) are updated using an exponential moving average (EMA), in which the parameters of model \( S \) are integrated into \( S' \), with the stop gradient (SG) operation applied. EMA and SG are applied during training to stabilize the learning process and reduce update variance. 
Although \( S' \) is updated from \( S \) via EMA, the two models retain complementary parameter subsets, which leads them to develop distinct yet structurally aligned feature representations. This differentiation reduces the likelihood of mutual error amplification. Moreover, the EMA update introduces temporal averaging, which stabilizes the supervisory signal by preventing abrupt changes in \( S' \), thereby functioning as a regularization mechanism rather than direct weight replication.
After each training iteration, the weights of the model $S'$ are updated using:
\begin{equation}
\theta_{S'} \leftarrow \beta \theta_{S'} + (1 - \beta) \theta_S.
\end{equation}
Here, $\beta$ is a smoothing factor that mitigates abrupt weight updates, thereby reducing the risk of overfitting and enhancing training stability. \par
SKD enables the model to leverage previously learned representations without relying on external teacher models, thereby supporting efficient and resource-aware training. After training, the final student model \( S \), refined through SKD, is used to perform image classification. Specifically, the learned weights of model \( S \) are utilized to classify images in the test dataset, and performance is evaluated using metrics such as classification accuracy and other relevant indicators. The proposed approach achieves efficient learning without incurring additional computational resources, making it well-suited for applications in resource-constrained medical settings.
\begin{table*}[t]
\centering
\caption{Experimental results of image classification using chest X-ray images. The 1\%, 5\%, and 10\% training data refer to randomly sampled subsets from the full training dataset. The best results are marked in bold.}
\begin{tabular}{cc@{\hskip 20pt}c@{\hskip 20pt}@{\hskip 20pt}c@{\hskip 20pt}@{\hskip 20pt}c@{\hskip 20pt}@{\hskip 20pt}c@{\hskip 20pt}}
\hline
Teacher model & Student model & Method & 1\% & 5\% & 10\% \\
\hline
\multirow{5}{*}{ViT-S} & \multirow{5}{*}{ViT-T} 
 & Ours & \textbf{0.750 $\pm$ 0.012} & \textbf{0.844 $\pm$ 0.004} & \textbf{0.869 $\pm$ 0.002} \\
 & & CM1 & 0.673 $\pm$ 0.006 & 0.791 $\pm$ 0.012 & 0.828 $\pm$ 0.002\\
 & & CM2 & 0.614 $\pm$ 0.023 & 0.639 $\pm$ 0.018 & 0.663 $\pm$ 0.009 \\
 & & CM3 & 0.628 $\pm$ 0.006 & 0.628 $\pm$ 0.009 & 0.667 $\pm$ 0.009 \\
 & & CM4 & 0.589 $\pm$ 0.007 & 0.747 $\pm$ 0.005 & 0.792 $\pm$ 0.021 \\
\hline
\multirow{5}{*}{ConvNeXt-T} & \multirow{5}{*}{ConvNeXt-F} 
 & Ours & \textbf{0.790 $\pm$ 0.002} & \textbf{0.885 $\pm$ 0.003} & \textbf{0.906 $\pm$ 0.004} \\
 & & CM1 & 0.751 $\pm$ 0.005 & 0.808 $\pm$ 0.012 & 0.850 
 $\pm$ 0.003 \\
 & & CM2 & 0.656 $\pm$ 0.013 & 0.756 $\pm$ 0.009 & 0.792 $\pm$ 0.006 \\
 & & CM3 & 0.656 $\pm$ 0.013 & 0.760 $\pm$ 0.010 & 0.794 $\pm$ 0.006 \\
 & & CM4 & 0.615 $\pm$ 0.016 & 0.738 $\pm$ 0.006 & 0.779 $\pm$ 0.007 \\
\hline
\end{tabular}
\label{tab2}
\end{table*}
\begin{figure*}[t]
    \centering
    \begin{minipage}{0.45\textwidth}
        \centering
        \includegraphics[width=\textwidth]{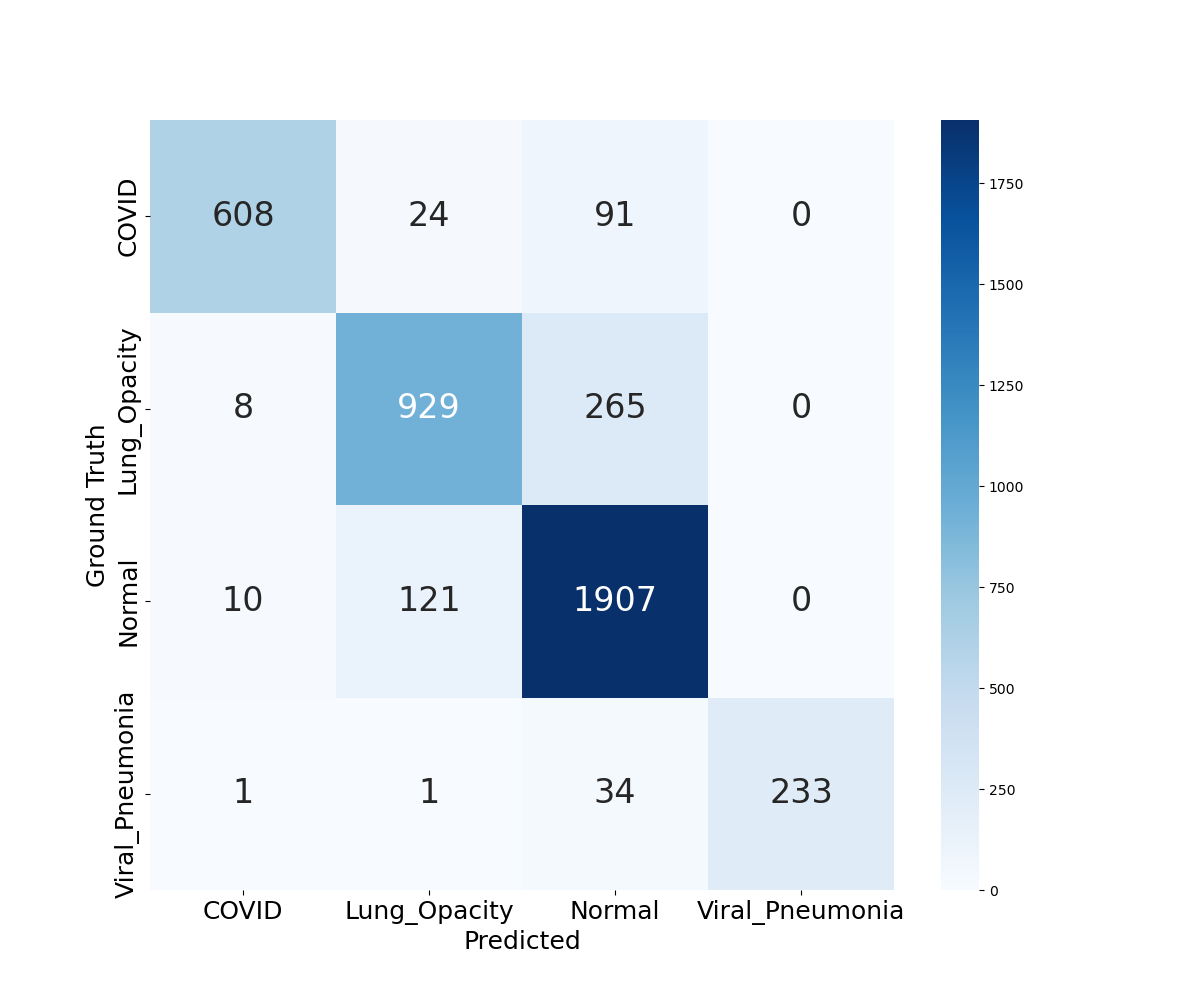}
        \subcaption{}
    \end{minipage}
    \begin{minipage}{0.45\textwidth}
        \centering
        \includegraphics[width=\textwidth]{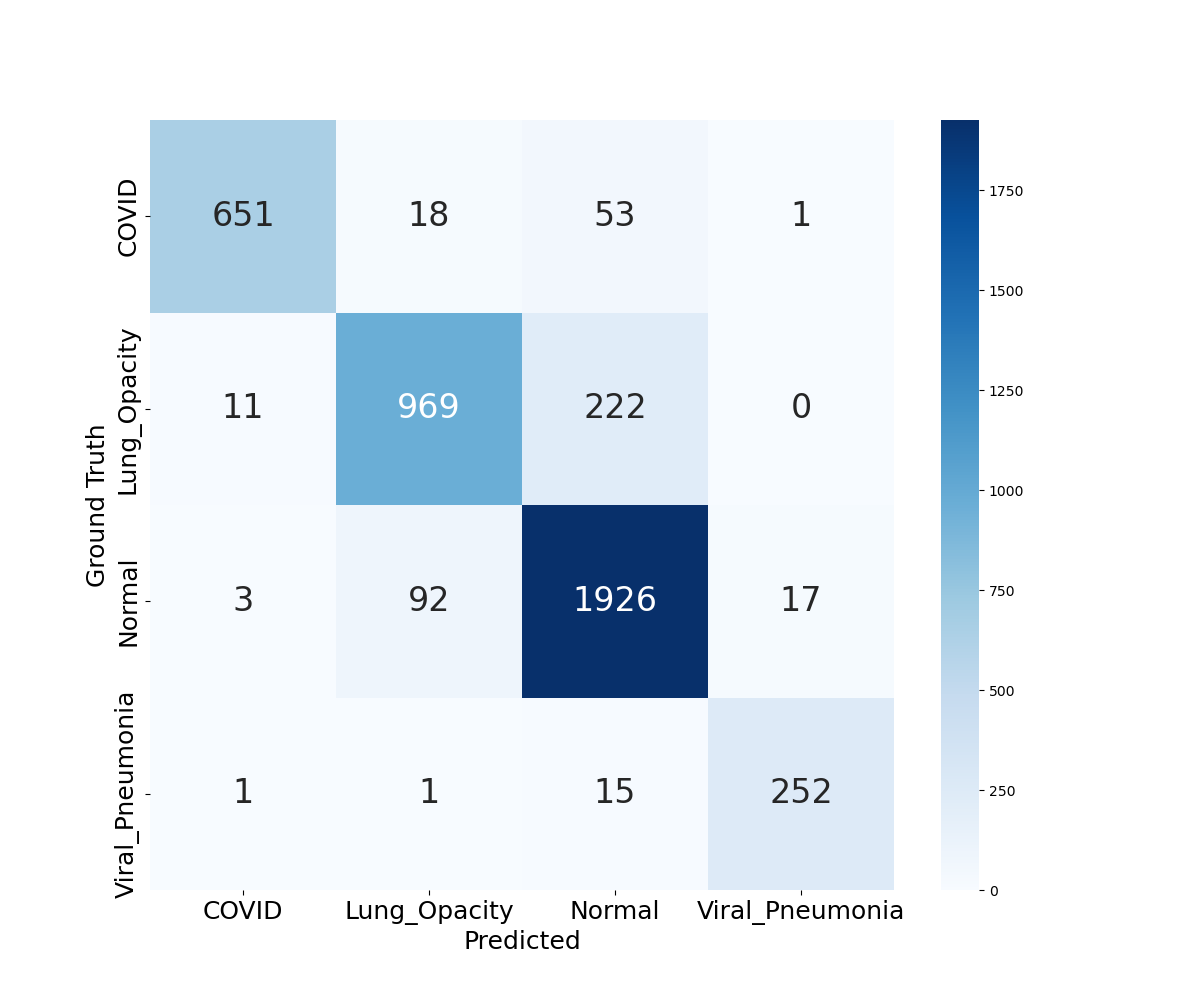}
        \subcaption{}
    \end{minipage}
    \caption{Confusion matrix of our method using 10\% training data of chest X-ray images. (a): ViT-T, (b): ConvNeXt-F.}
    \label{cf1}
\end{figure*}
\begin{table*}[t]
\centering
\caption{Experimental results of image classification using lung CT images. The 1\%, 5\%, and 10\% training data refer to randomly sampled subsets from the full training dataset. The best results are marked in bold.}
\begin{tabular}{cc@{\hskip 20pt}c@{\hskip 20pt}@{\hskip 20pt}c@{\hskip 20pt}@{\hskip 20pt}c@{\hskip 20pt}@{\hskip 20pt}c@{\hskip 20pt}}
\hline
Teacher model & Student model & Method & 1\% & 5\% & 10\% \\
\hline
\multirow{5}{*}{ViT-S} & \multirow{5}{*}{ViT-T} 
 & Ours & \textbf{0.736 $\pm$ 0.007} & \textbf{0.828 $\pm$ 0.005} & \textbf{0.889 $\pm$ 0.003} \\
 & & CM1 & 0.694 $\pm$ 0.007 & 0.806 $\pm$ 0.007 & 0.870 $\pm$ 0.006\\
 & & CM2 & 0.683 $\pm$ 0.006 & 0.757 $\pm$ 0.012 & 0.743 $\pm$ 0.014 \\
 & & CM3 & 0.679 $\pm$ 0.003 & 0.757 $\pm$ 0.013 & 0.807 $\pm$ 0.028 \\
 & & CM4 &  0.680 $\pm$ 0.003  &  0.729 $\pm$ 0.022  &  0.779 $\pm$ 0.041  \\
\hline
\multirow{5}{*}{ConvNeXt-T} & \multirow{5}{*}{ConvNeXt-F} 
 & Ours & \textbf{0.777 $\pm$ 0.017} & \textbf{0.884 $\pm$ 0.003} &  \textbf{0.925 $\pm$ 0.005} \\
 & & CM1 & 0.700 $\pm$ 0.007 & 0.853 $\pm$ 0.009 & 0.904 
 $\pm$ 0.005 \\
 & & CM2 & 0.660 $\pm$ 0.003 & 0.806 $\pm$ 0.009 & 0.875 $\pm$ 0.006 \\
 & & CM3 & 0.638 $\pm$ 0.014 & 0.799 $\pm$ 0.006 & 0.866 $\pm$ 0.007 \\
 & & CM4 & 0.668 $\pm$ 0.008 & 0.800 $\pm$ 0.006 & 0.868 $\pm$ 0.005 \\
 \hline
\end{tabular}
\label{tab3}
\end{table*}
\begin{figure*}[t]
    \centering
    \begin{minipage}{0.45\textwidth}
        \centering
        \includegraphics[width=\textwidth]{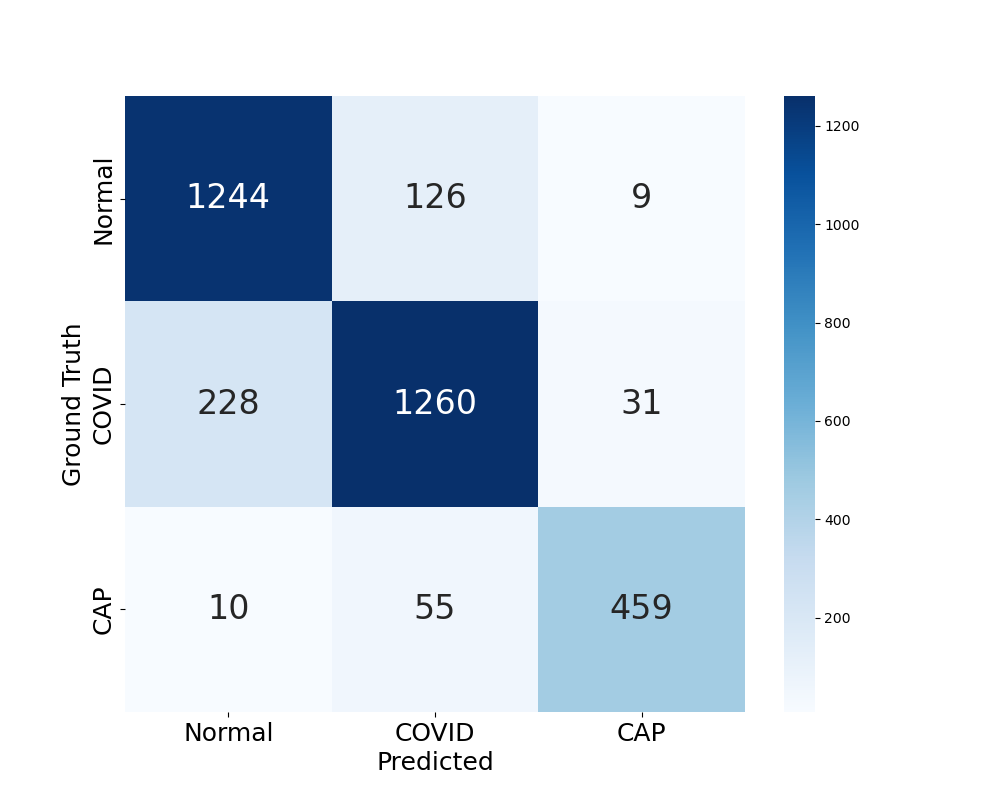}
        \subcaption{}
    \end{minipage}
    \begin{minipage}{0.45\textwidth}
        \centering
        \includegraphics[width=\textwidth]{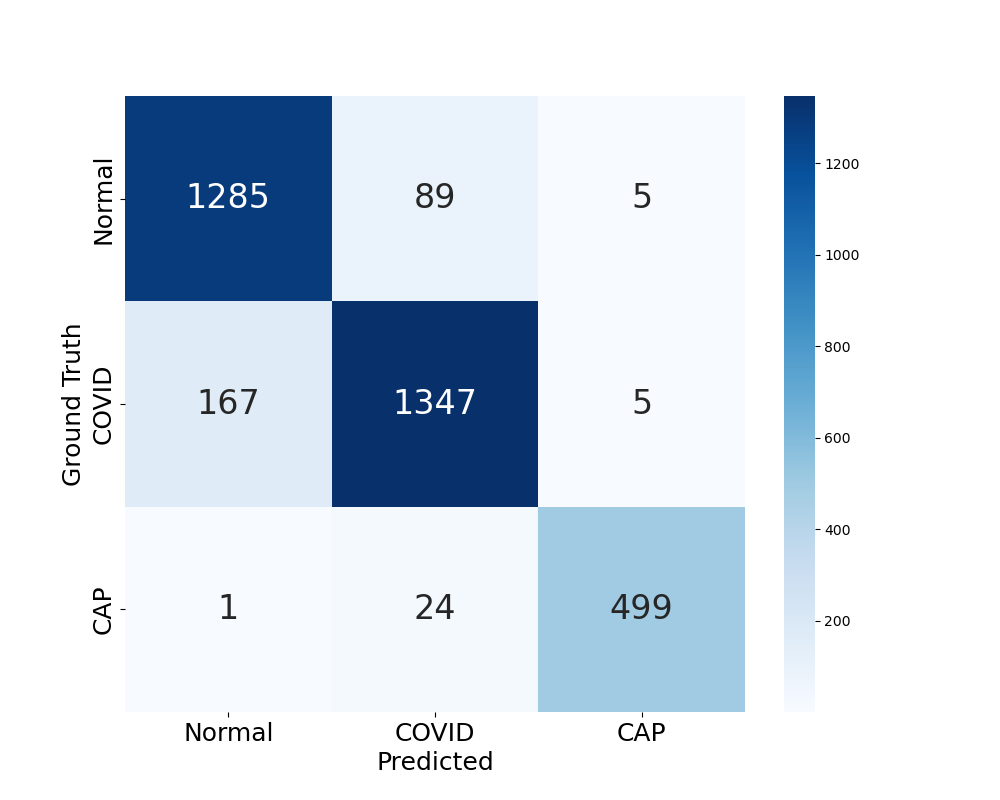}
        \subcaption{}
    \end{minipage}
    \caption{Confusion matrix of our method using 10\% training data of lung CT images. (a): ViT-T, (b): ConvNeXt-F.}
    \label{cf2}
\end{figure*}
\begin{table*}[t]
\centering
\caption{Experimental results of image classification using brain MRI images. The 1\%, 5\%, and 10\% training data refer to randomly sampled subsets from the full training dataset. The best results are marked in bold.}
\begin{tabular}{cc@{\hskip 20pt}c@{\hskip 20pt}@{\hskip 20pt}c@{\hskip 20pt}@{\hskip 20pt}c@{\hskip 20pt}@{\hskip 20pt}c@{\hskip 20pt}}
\hline
Teacher model & Student model & Method & 1\% & 5\% & 10\% \\
\hline
 \multirow{5}{*}{ViT-S} & \multirow{5}{*}{ViT-T} 
 & Ours & \textbf{0.665 $\pm$ 0.034} & \textbf{0.774 $\pm$ 0.019} & \textbf{0.848 $\pm$ 0.016} \\
 & & CM1 & 0.663 $\pm$ 0.025 & 0.766 $\pm$ 0.014 & 0.839 $\pm$ 0.012\\
 & & CM2 & 0.617 $\pm$ 0.004 & 0.741 $\pm$ 0.010 & 0.803 $\pm$ 0.006 \\
 & & CM3 & 0.602 $\pm$ 0.004 & 0.710 $\pm$ 0.007  &  0.771 $\pm$ 0.015  \\
 & & CM4 & 0.584 $\pm$ 0.007 & 0.700 $\pm$ 0.009 & 0.772 $\pm$ 0.016 \\
\hline
\multirow{5}{*}{ConvNeXt-T} & \multirow{5}{*}{ConvNeXt-F} 
 & Ours & \textbf{0.747 $\pm$ 0.029} & \textbf{0.852 $\pm$ 0.012} & \textbf{0.910 $\pm$ 0.014} \\
 & & CM1 & 0.656 $\pm$ 0.027 & 0.766 $\pm$ 0.017 & 0.837 $\pm$ 0.013 \\
 & & CM2 & 0.581 $\pm$ 0.016 & 0.680 $\pm$ 0.009 & 0.758 $\pm$ 0.012 \\
 & & CM3 & 0.586 $\pm$ 0.015 & 0.682 $\pm$ 0.008 & 0.758 $\pm$ 0.010 \\
 & & CM4 & 0.586 $\pm$ 0.015 & 0.681 $\pm$ 0.009 & 0.757 $\pm$ 0.010 \\
\hline
\end{tabular}
\label{tab4}
\end{table*}
\begin{figure*}[t]
    \centering
    \begin{minipage}{0.45\textwidth}
        \centering
        \includegraphics[width=\textwidth]{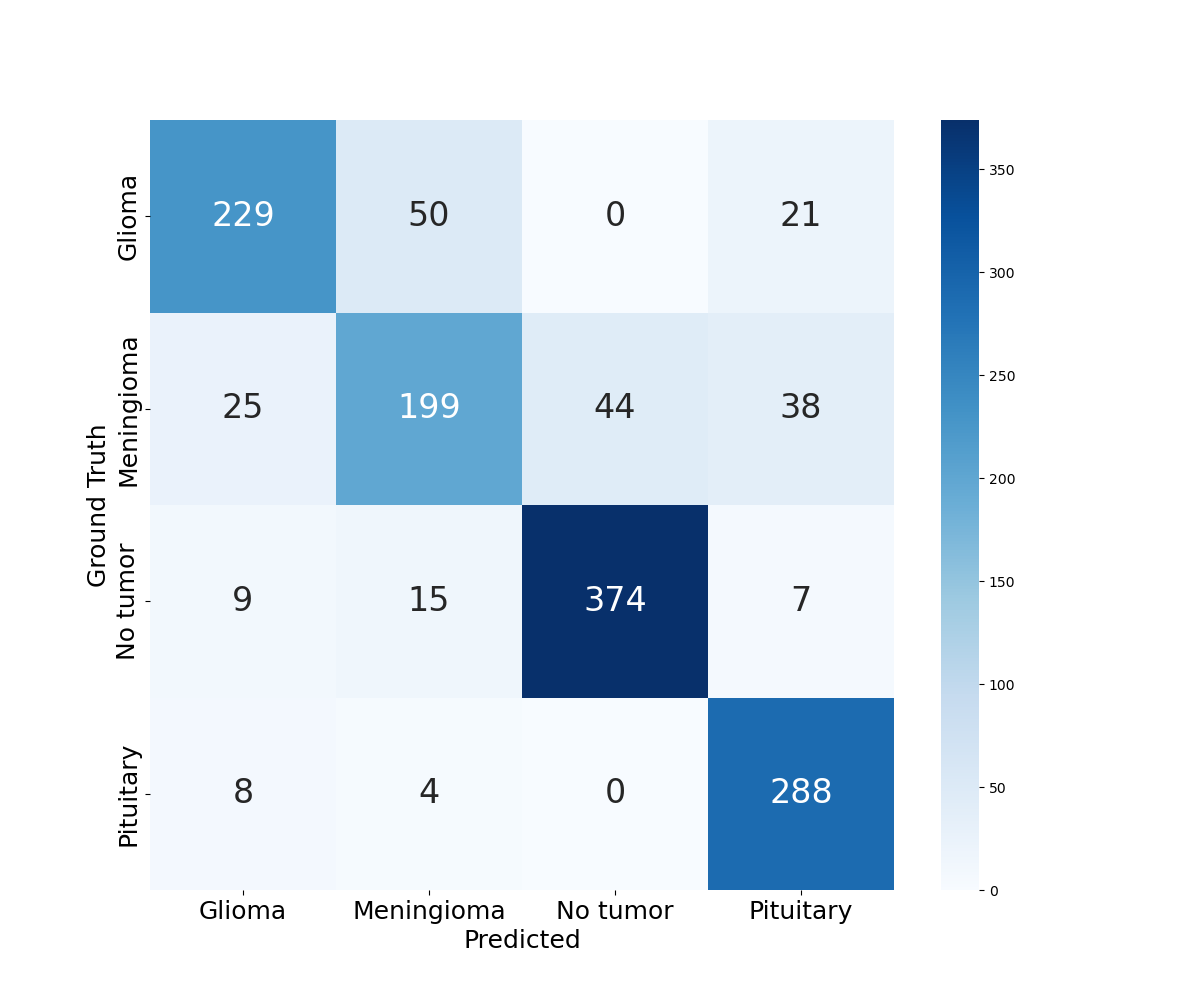}
        \subcaption{}
    \end{minipage}
    \begin{minipage}{0.45\textwidth}
        \centering
        \includegraphics[width=\textwidth]{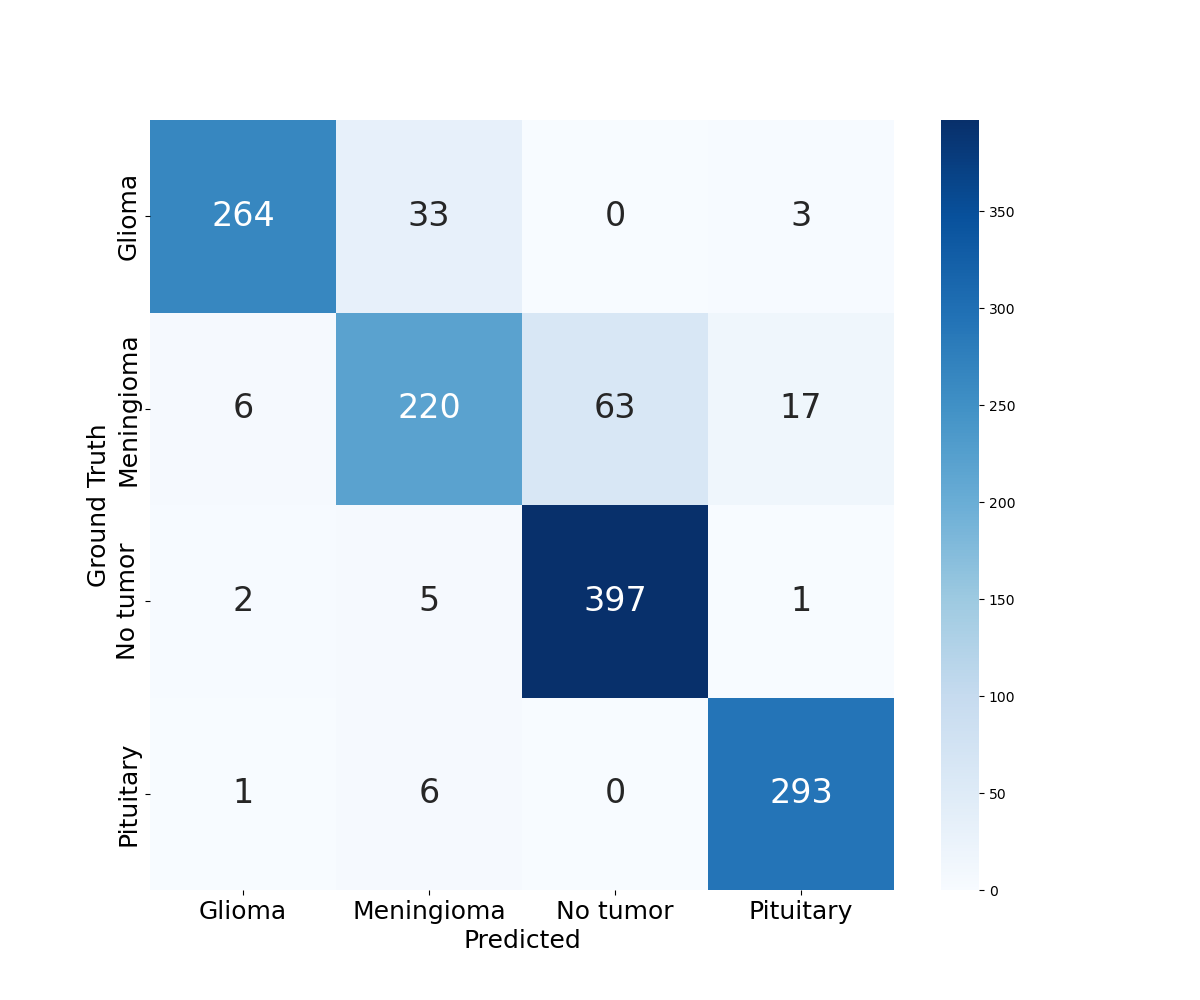}
        \subcaption{}
    \end{minipage}
    \caption{Confusion matrix of our method using 10\% training data of brain MRI images. (a): ViT-T, (b): ConvNeXt-F.}
    \label{cf3}
\end{figure*}
\begin{table}[t]
\centering
\caption{Experimental results of image classification on three full datasets.}
\begin{tabular}{cccc}
\hline
 Model & Dataset & Ours & CM1 \\
\hline
 \multirow{3}{*}{ViT} & X-ray & 0.990 & 0.949 \\
 & CT & 0.994 & 0.992 \\
 & MRI & 0.992 & 0.989 \\ \hline
 \multirow{3}{*}{ConvNeXt} & X-ray & 0.956 & 0.942 \\
 & CT & 0.995 & 0.994 \\
 & MRI & 0.998 & 0.996 \\ \hline
\end{tabular}
\label{tab9}
\end{table}
\begin{table}[t]
\centering
\caption{Experimental results of image classification on three extremely limited (0.5\%) training datasets.}
\begin{tabular}{cccc}
\hline
 Model & Dataset & Ours & CM1 \\
\hline
 \multirow{3}{*}{ViT} & X-ray & 0.679 $\pm$ 0.012 & 0.672 $\pm$ 0.015 \\
 & CT & 0.695 $\pm$ 0.010 & 0.692 $\pm$ 0.011 \\
 & MRI & 0.610 $\pm$ 0.027 & 0.602 $\pm$ 0.028 \\ \hline
\end{tabular}
\label{tab11}
\end{table}
\begin{table*}[t]
\centering
\caption{Evaluation using other metrics with a 10\% training data of each dataset.}
\begin{tabular}{cccccccc}
\hline
  \multirow{2}{*}{Model} & \multirow{2}{*}{Dataset} & \multicolumn{3}{c}{Ours} & \multicolumn{3}{c}{CM1} \\
  &  & ACC & F1 Score & AUC & ACC & F1 Score & AUC \\
 \hline
 \multirow{3}{*}{ViT} & X-ray & 0.866 $\pm$ 0.003 & 0.878 $\pm$ 0.003 & 0.968 $\pm$ 0.001 & 0.839 $\pm$ 0.006 & 0.842 $\pm$ 0.008 & 0.958 $\pm$ 0.002 \\
 & CT & 0.897 $\pm$ 0.008 & 0.884 $\pm$ 0.010 & 0.962 $\pm$ 0.003 & 0.843 $\pm$ 0.010 & 0.846 $\pm$ 0.011 & 0.943 $\pm$ 0.003 \\
 & MRI & 0.844 $\pm$ 0.012 & 0.837 $\pm$ 0.014 & 0.964 $\pm$ 0.003 & 0.828 $\pm$ 0.007 & 0.818 $\pm$ 0.008 & 0.959 $\pm$ 0.004 \\ \hline
 \multirow{3}{*}{ConvNeXt} & X-ray & 0.909 $\pm$ 0.003 & 0.920 $\pm$ 0.002 & 0.982 $\pm$ 0.001 & 0.838 $\pm$ 0.011 & 0.839 $\pm$ 0.014 & 0.962 $\pm$ 0.017 \\
 & CT & 0.925 $\pm$ 0.004 & 0.931 $\pm$ 0.004 & 0.981 $\pm$ 0.003 & 0.906 $\pm$ 0.002 & 0.917 $\pm$ 0.002 & 0.972 $\pm$ 0.001 \\
 & MRI & 0.900 $\pm$ 0.015 & 0.895 $\pm$ 0.015 & 0.980 $\pm$ 0.005 & 0.835 $\pm$ 0.010 & 0.827 $\pm$ 0.010 & 0.958 $\pm$ 0.004 \\ \hline
\end{tabular}
\label{tab10}
\end{table*}
\section{Experiments}
\label{sec4}
This section presents extensive experiments and hyperparameter analyses to evaluate the effectiveness of the proposed medical image classification approach.
Details of the datasets and experimental settings are provided in Subsection 4.1.
The main results are shown in Subsection 4.2.
An analysis of model performance when interchanging the roles of the main and auxiliary models is given in Subsection 4.3.
Hyperparameter sensitivity analysis is reported in Subsection 4.4.
Comparative analysis of different SKD strategies is provided in Subsection 4.5.
The analysis of weight selection methods is shown in Subsection 4.6.
Computational cost and training time analysis are presented in Subsection 4.7.
\subsection{Dataset and Settings}
\label{4.1}
We conducted experiments using publicly available datasets of chest X-ray, lung CT scan, and brain MRI images. The chest X-ray dataset \footnote{\url{https://www.kaggle.com/datasets/tawsifurrahman/covid19-radiography-database}} contains 21,165 images classified into four categories: COVID-19, lung opacity, normal, and viral pneumonia. The lung CT dataset \footnote{\url{https://www.kaggle.com/datasets/maedemaftouni/large-covid19-ct-slice-dataset}} comprises 17,104 images classified into three categories: COVID-19, normal, and community-acquired pneumonia (CAP). The brain MRI dataset \footnote{\url{https://www.kaggle.com/datasets/masoudnickparvar/brain-tumor-mri-dataset}} includes 7,023 images divided into four classes: glioma, meningioma, no tumor, and pituitary. 
For each dataset, 80\% of the data was randomly selected for training, and the remaining 20\% was reserved for testing. All images were resized to 224 $\times$ 224 pixels. Classification accuracy was used to evaluate performance on the four-class chest X-ray and brain MRI datasets, as well as the three-class lung CT datasets. We evaluated our method on the test sets of all three datasets to assess generalizability. 
To examine performance under limited data conditions, we randomly selected 1\%, 5\%, and 10\% of the training data, repeating each sampling
five times. \par 
ViT-S~\cite{dosovitskiy2021image} and ConvNeXt-T~\cite{liu2022convnet} pretrained on ImageNet-21K~\cite{deng2009imagenet} were used as teacher models, while ViT-T and ConvNeXt-F served as student models. Pretraining on the large ImageNet-21K dataset enables the models to learn more effectively than training solely on small medical image datasets. Table~\ref{tab1} summarizes the architectural details of the teacher and student models. Model depth indicates the total number of layers, reflecting the capacity to learn hierarchical features. The embedding dimension denotes the size of feature vectors at each layer, which determines the representational power of each token or feature map. In ConvNeXt architectures, both depth and embedding dimension vary across stages, indicating progressively increasing model complexity. For example, ConvNeXt-T uses a stage-wise depth of 3 / 3 / 9 / 3 with corresponding embedding dimensions of 96 / 192 / 384 / 768.

Comparative experiments included the following methods. CM1, weight selection applied to a single student model (without SKD)~\cite{xu2024initializing};
CM2, default initialization from the timm library (without weight selection)~\cite{rw2019timm}; CM3, Xavier initialization~\cite{pmlr-v9-glorot10a}; and CM4, Kaiming initialization~\cite{he2015delving}. 
For ViT experiments, training ran for 300 epochs with the temperature parameter $\tau$ set to 4, $\alpha$ to 0.6, and $\beta$ to 0.9.
For ConvNeXt experiments, training also lasted 300 epochs, with $\tau$ set to 3, $\alpha$ to 0.6, and $\beta$ to 0.9.
Detailed hyperparameter analyses are presented in Subsection 4.4.
All experiments were conducted on an NVIDIA RTX A6000 GPU using the PyTorch framework.
\begin{table*}[t]
\centering
\caption{Experimental results with the main and auxiliary models interchanged. The 1\%, 5\%, and 10\% training data refer to randomly sampled subsets from the entire training dataset. Ours uses the original configuration, while Ours\_Swap swaps the roles of the main and auxiliary models.}
\begin{tabular}{@{\hskip 22pt}c@{\hskip22pt}c@{\hskip 22pt}c@{\hskip 22pt}@{\hskip 22pt}c@{\hskip 22pt}@{\hskip 22pt}c@{\hskip 22pt}@{\hskip 22pt}c@{\hskip 22pt}}
\hline
 Model & \multicolumn{2}{c}{Dataset} & Ours & Ours\_Swap & CM1 \\
\hline
 \multirow{9}{*}{ViT} & \multirow{3}{*}{X-ray}& 1\% & 0.750 $\pm$ 0.012 & 0.742 $\pm$ 0.013 & 0.673 $\pm$ 0.006 \\
 & & 5\% & 0.844 $\pm$ 0.004 & 0.841 $\pm$ 0.003 & 0.791 $\pm$ 0.012 \\
 & & 10\% & 0.869 $\pm$ 0.002 & 0.871 $\pm$ 0.001 & 0.828 $\pm$ 0.002 \\ \cline{2-6}
 & \multirow{3}{*}{CT} & 1\% & 0.736 $\pm$ 0.007 & 0.728 $\pm$ 0.010 & 0.694 $\pm$ 0.007 \\
 & & 5\% & 0.828 $\pm$ 0.005 & 0.831 $\pm$ 0.005 & 0.806 $\pm$ 0.007 \\
 & & 10\% & 0.889 $\pm$ 0.003 & 0.888 $\pm$ 0.004 & 0.870 $\pm$ 0.006 \\ \cline{2-6}
 & \multirow{3}{*}{MRI} & 1\% & 0.665 $\pm$ 0.030 & 0.664 $\pm$ 0.031 & 0.663 $\pm$ 0.025 \\
 & & 5\% & 0.774 $\pm$ 0.019 & 0.777 $\pm$ 0.015 & 0.766 $\pm$ 0.014 \\
 & & 10\% & 0.848 $\pm$ 0.016 & 0.841 $\pm$ 0.001 & 0.839 $\pm$ 0.012 \\
\hline
 \multirow{9}{*}{ConvNeXt} & \multirow{3}{*}{X-ray} & 1\% & 0.790 $\pm$ 0.002 & 0.791 $\pm$ 0.002 & 0.751 $\pm$ 0.005 \\
 &  & 5\% & 0.885 $\pm$ 0.003 & 0.884 $\pm$ 0.004 & 0.808 $\pm$ 0.012 \\
 & & 10\% & 0.906 $\pm$ 0.004 & 0.907 $\pm$ 0.004 & 0.850 $\pm$ 0.003 \\ \cline{2-6}
 & \multirow{3}{*}{CT} & 1\% & 0.777 $\pm$ 0.017 & 0.787 $\pm$ 0.013 & 0.700 $\pm$ 0.007 \\
 & & 5\% & 0.884 $\pm$ 0.003 & 0.885 $\pm$ 0.004 & 0.853 $\pm$ 0.009 \\
 & & 10\% & 0.925 $\pm$ 0.005 & 0.923 $\pm$ 0.005 & 0.904 $\pm$ 0.005 \\ \cline{2-6}
 & \multirow{3}{*}{MRI} & 1\% & 0.747 $\pm$ 0.029 &  0.754 $\pm$ 0.023 & 0.656 $\pm$ 0.027 \\
 & & 5\% & 0.852 $\pm$ 0.012 & 0.859 $\pm$ 0.011 & 0.766 $\pm$ 0.017 \\
 & & 10\% & 0.910 $\pm$ 0.014 & 0.909 $\pm$ 0.015 & 0.837 $\pm$ 0.013 \\
\hline
\end{tabular}
\label{tab6}
\end{table*}
\begin{figure*}[t]
    \centering
    \begin{minipage}{0.3\textwidth}
        \centering
        \includegraphics[width=\textwidth]{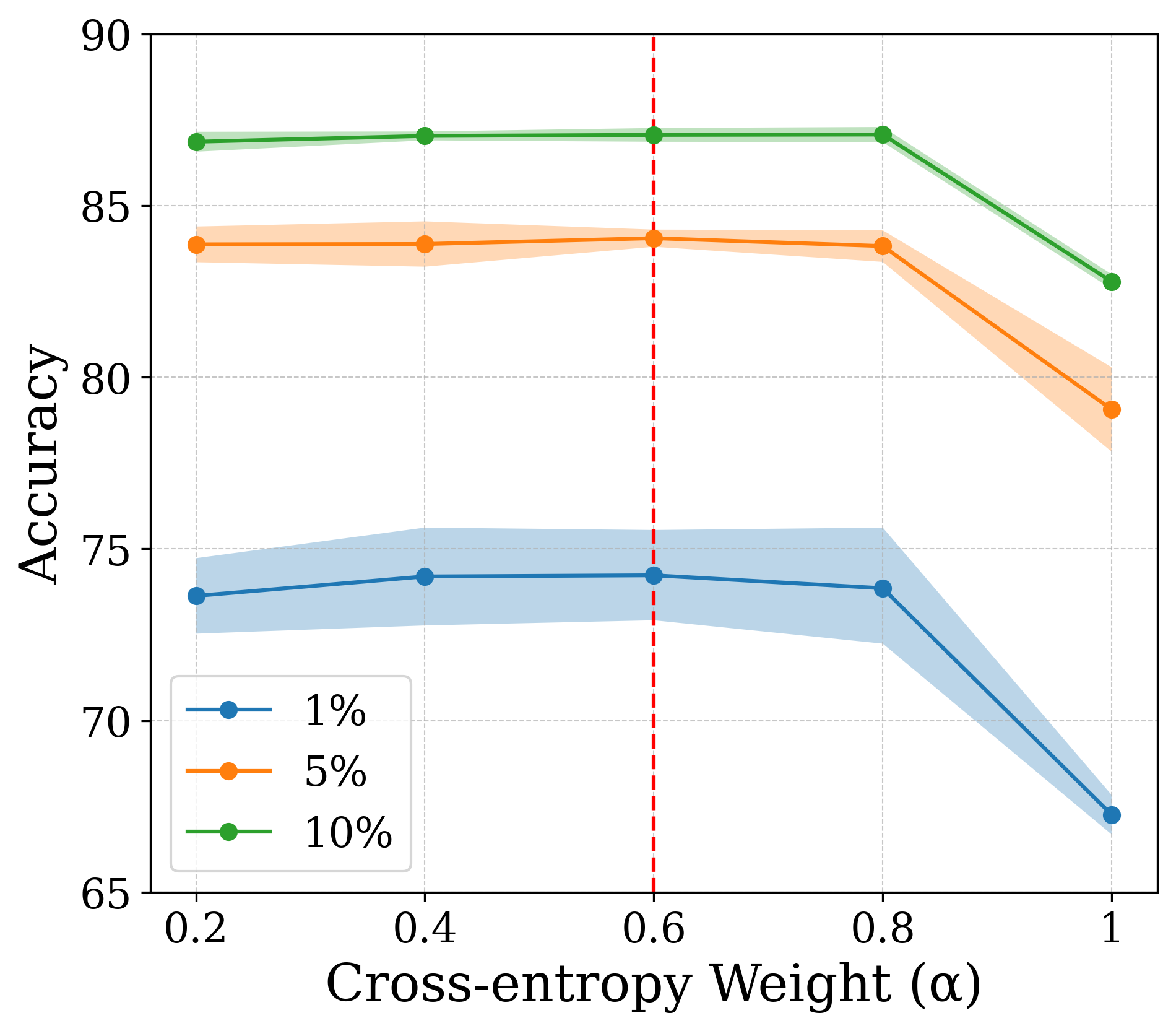}
        \subcaption{}
    \end{minipage}
    \hfill
    \begin{minipage}{0.3\textwidth}
        \centering
        \includegraphics[width=\textwidth]{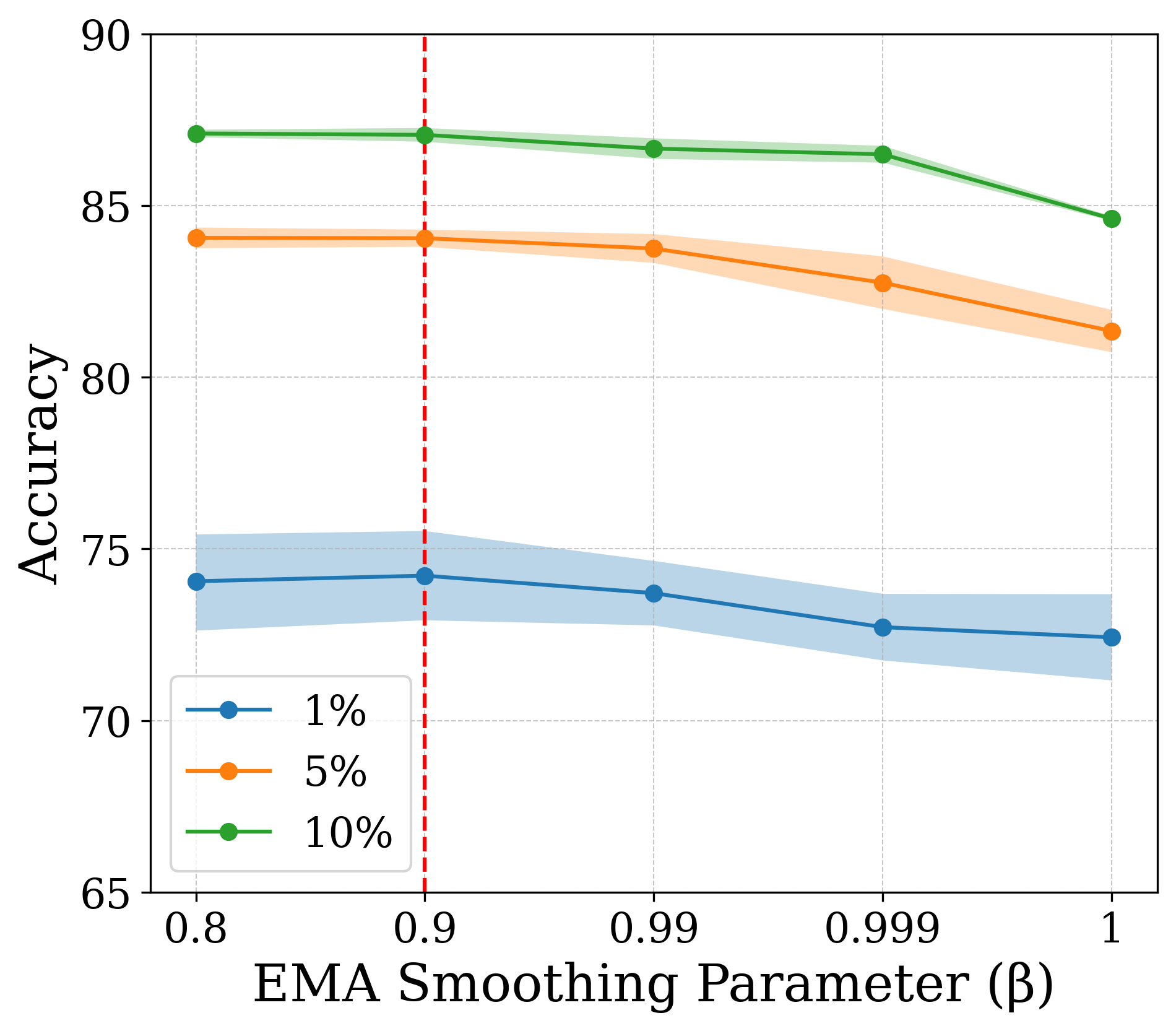}
        \subcaption{}
    \end{minipage}
    \hfill
    \begin{minipage}{0.3\textwidth}
        \centering
        \includegraphics[width=\textwidth]{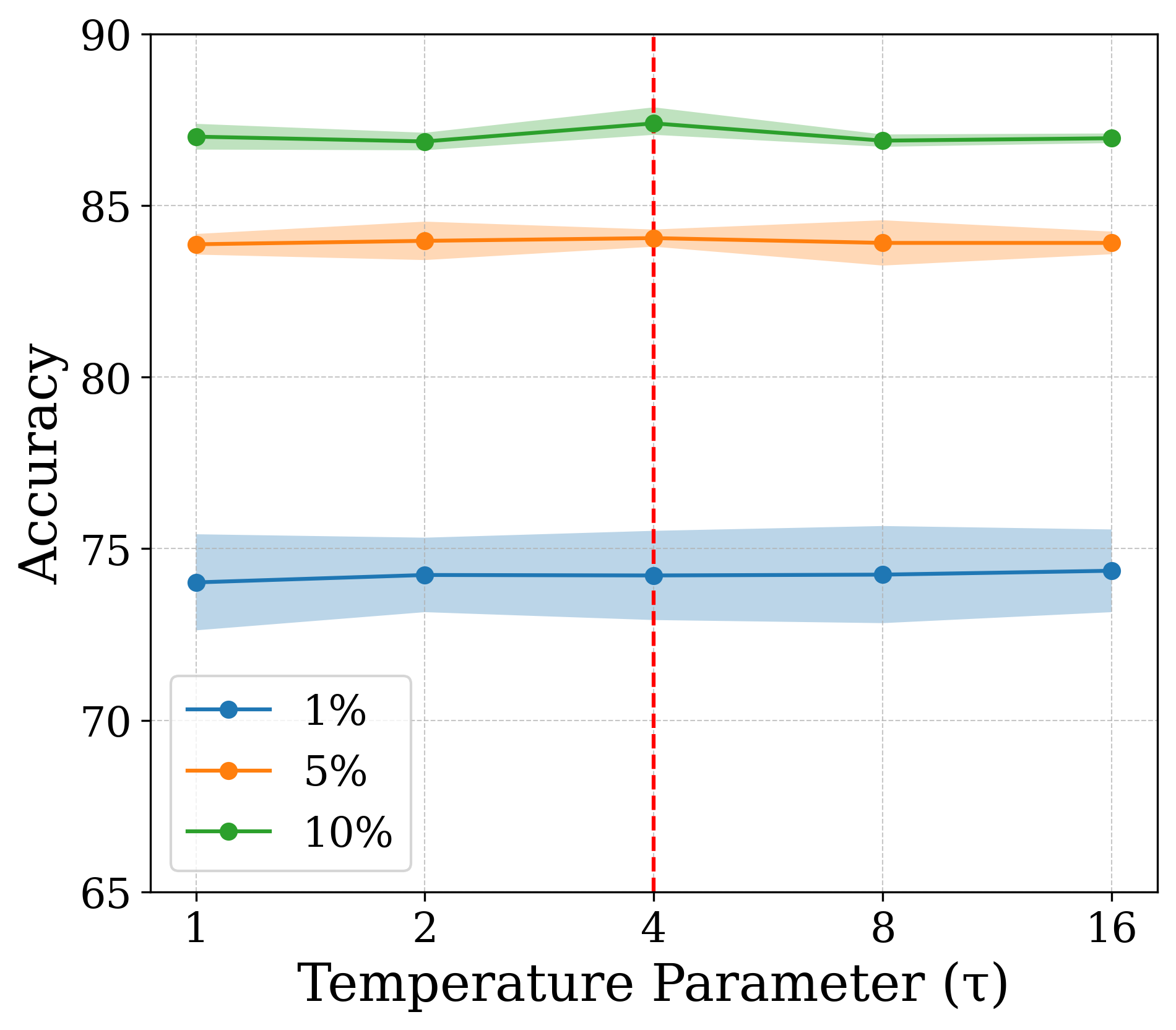}
        \subcaption{}
    \end{minipage}
    \caption{The analysis of the hyperparameters: the cross-entropy weight ($\alpha$), the EMA smoothing parameter ($\beta$), and the temperature parameter ($\tau$). The results were obtained using ViT-S as the teacher model and ViT-T as the student model, with different subsets. The 1\%, 5\%, and 10\% training data refer to the randomly sampled subsets of the entire training dataset. Each configuration was tested five times, with the points representing the average values, and the shaded areas indicating the range between the minimum and maximum accuracy. Dashed lines indicate the performance of the settings adopted in Subsection 4.2.}
    \label{fig3}
\end{figure*}
\begin{table*}[t]
\centering
\caption{Experimental results using various self-knowledge distillation methods with weight selection. The 1\%, 5\%, and 10\% training data refer to randomly sampled subsets of the entire training dataset.}
\begin{tabular}{@{\hskip 30pt}c@{\hskip30pt}l@{\hskip 35pt}c@{\hskip 25pt}@{\hskip 25pt}c@{\hskip 25pt}@{\hskip 25pt}c@{\hskip 25pt}}
\hline
Dataset & Method & 1\% & 5\% & 10\% \\
\hline
\multirow{4}{*}{X-ray} 
 & Ours (SKD) & 0.750 $\pm$ 0.012 & 0.844 $\pm$ 0.004 & 0.869 $\pm$ 0.002\\
 & Ours (CS-KD) & 0.727 $\pm$ 0.013 & 0.830 $\pm$ 0.005 & 0.869 $\pm$ 0.003\\
 & Ours (TF-KD) & 0.725 $\pm$ 0.017 & 0.844 $\pm$ 0.004 & 0.873 $\pm$ 0.003 \\
 & CM1 & 0.673 $\pm$ 0.006 & 0.791 $\pm$ 0.012 & 0.828 $\pm$ 0.002 \\
\hline
\multirow{4}{*}{CT} 
 & Ours (SKD) & 0.736 $\pm$ 0.007 & 0.828 $\pm$ 0.005 & 0.889 $\pm$ 0.003 \\
 & Ours (CS-KD) & 0.740 $\pm$ 0.015 & 0.835 $\pm$ 0.007 & 0.889 $\pm$ 0.003\\
 & Ours (TF-KD) & 0.714 $\pm$ 0.004 & 0.828 $\pm$ 0.006 & 0.887 $\pm$ 0.007 \\
 & CM1 & 0.694 $\pm$ 0.007 & 0.806 $\pm$ 0.007 & 0.870 $\pm$ 0.006 \\
\hline
\multirow{4}{*}{MRI} 
 & Ours (SKD) & 0.665 $\pm$ 0.030 & 0.774 $\pm$ 0.019 & 0.848 $\pm$ 0.016 \\
 & Ours (CS-KD) & 0.666 $\pm$ 0.034 & 0.792 $\pm$ 0.012 & 0.849 $\pm$ 0.008\\
 & Ours (TF-KD) & 0.658 $\pm$ 0.030 & 0.781 $\pm$ 0.016 & 0.852 $\pm$ 0.013 \\
 & CM1 & 0.663 $\pm$ 0.025 & 0.766 $\pm$ 0.014 & 0.839 $\pm$ 0.008 \\
\hline
\end{tabular}
\label{tab5}
\end{table*}
\subsection{Main Results}
\label{4.2}
Tables~\ref{tab2}, \ref{tab3}, and \ref{tab4} present the classification results for chest X-ray, lung CT scan, and brain MRI images, respectively. Each experiment was repeated five times, and both the mean and standard deviation were recorded. For each dataset, training subsets corresponding to 1\%, 5\%, and 10\% of the full dataset were constructed. To minimize sampling bias and maintain comparability, stratified random sampling was employed to preserve the original class distribution. The sampling procedure was repeated five times with distinct random seeds, yielding five independent experimental splits for each subset ratio. All models were trained separately on these splits, and the mean and standard deviation of the performance metrics are reported, ensuring a robust evaluation of the method’s stability under different data availability conditions.
These tables compare the performance of CM1, CM2, CM3, and CM4 with the proposed method, which combines weight selection and SKD (Ours). \par
Our method consistently outperformed all comparison methods in classification accuracy across all datasets, as shown in Tables \ref{tab2}, \ref{tab3}, and \ref{tab4}. 
By integrating dual-model weight selection with SKD, it enables efficient knowledge transfer and reuse, significantly improving feature retention
and learning efficiency. Compared to CM1, which applies weight selection alone, our approach significantly improves feature retention and learning efficiency. 
Moreover, by evaluating different initialization strategies, we observed that our method substantially enhances neural network initialization, yielding more effective training than CM2, CM3, and CM4. 
The consistent superiority across datasets demonstrates the robustness and generalizability of our approach, even in low-data regimes. Notably, the significant improvements over CM2–CM4 emphasize that default or classical initialization schemes alone are insufficient for effective learning in smaller models. Our method not only boosts accuracy but also ensures stability across trials, as reflected in the low standard deviations reported. 
Figures~\ref{cf1}, \ref{cf2}, and \ref{cf3} illustrate the confusion matrices of our method on the chest X-ray, lung CT scan, and brain MRI datasets, respectively. The results demonstrate consistently high accuracy and robust generalization across all modalities. In the chest X-ray dataset (Fig.~\ref{cf1}), the model accurately distinguishes COVID-19 from other classes, between clinically similar conditions such as viral pneumonia and lung opacity, underscoring its robustness even with limited data. In the lung CT dataset (Fig.~\ref{cf2}), the model shows accurate classification, especially between COVID-19 and normal cases, which are often difficult to separate due to radiological differences. For the brain MRI dataset (Fig.~\ref{cf3}), the model effectively identifies tumor types such as glioma and meningioma, while reducing misclassification involving the no tumor class. These results confirm that combining dual-model weight selection with SKD enhances feature representation and discrimination, even when trained on just 10\% of the available data. \par
Furthermore, additional experiments were conducted using the full dataset to provide a more complete comparison with conventional fine-tuning. As shown in Table~\ref{tab9}, results from the full dataset confirmed the effectiveness of this method even with large-scale data. In particular, our method achieved higher classification accuracy in the chest X-ray dataset, which includes a larger number of images and more classes, compared to other datasets, demonstrating a strong generalization capability for complex data. On the other hand, since the performance improvement was more pronounced in the limited-data setting, the proposed method is especially effective when training data are scarce.
\par
Additionally, Table~\ref{tab11} summarizes the results obtained using only 0.5\% of the training data. These findings further highlight the advantage of our method in extremely data-limited scenarios. Even with such a minimal amount of training samples, the proposed approach consistently outperformed conventional fine-tuning methods, demonstrating strong robustness against data scarcity. This result reinforces that the proposed framework is particularly suitable for medical imaging applications, where annotated data are often minimal.
To ensure a more thorough evaluation, we supplemented accuracy with the F1-score and AUC, particularly given the class imbalance present in medical imaging datasets. These metrics were computed under the same experimental protocol applied in Figures~\ref{cf1}–\ref{cf3}.
As shown in Table~\ref{tab10}, our method achieves superior performance across all three metrics, demonstrating consistent advantages in both classification effectiveness and decision boundary quality.
\subsection{Analysis of Interchanged Main and Auxiliary Models}
\label{4.3}
In the proposed method, different subsets of weights from a single teacher model are transferred to two student models, followed by SKD. In this setup, one student serves as the primary (main) model, while the other functions as the auxiliary model providing soft targets. In this section, we compare the classification performance when the roles of the main and auxiliary models are reversed. The results of this experiment are presented in Table~\ref{tab6}. We denote the original configuration, where the main and auxiliary models are fixed, as Ours, and the configuration in which their roles are swapped as Ours\_Swap. The experimental results show that Ours\_Swap achieves performance comparable to Ours, confirming the robustness and reliability of the proposed method.

These results suggest that the proposed method is robust and not sensitive to model role assignment, and not dependent on which model is designated as the main or auxiliary. Because both student models are initialized with distinct yet complementary weight subsets from the same teacher model, each can support effective training through SKD. This means that our approach works reliably even if the roles of the models are swapped. Such flexibility is particularly beneficial in real-world deployment scenarios, where rigid role assignments may be impractical. It allows for more adaptable training pipelines without extensive manual tuning.
\subsection{Analysis of Hyperparameters}
\label{4.4}

We conducted experiments using ViT-S as the teacher model and ViT-T as the student model to investigate the influence of three key hyperparameters: $\alpha$, $\beta$, and $\tau$. 
The results are summarized in Fig.~\ref{fig3}. 
When $\alpha = 1$, the model exhibited performance comparable to training without SKD, since the distillation loss $L_{\text{skd}}$ was effectively excluded. 
As $\alpha$ decreased below 1, the inclusion of $L_{\text{skd}}$ consistently improved accuracy, demonstrating the effectiveness of self-knowledge distillation. 
For the EMA momentum $\beta$, the highest performance was achieved around $\beta = 0.9$, whereas larger values slightly degraded accuracy due to over-smoothing, which can limit the model’s adaptability to new gradient updates. 
Regarding the temperature parameter $\tau$, classification accuracy remained relatively stable across a broad range of values, suggesting that the model is insensitive to $\tau$ once sufficient training convergence is reached.

\subsection{Analysis of Self-knowledge Distillation Methods}
\label{4.5}
\begin{table*}[t]
\centering
\caption{Experimental results with various weight selection methods. 1\%, 5\%, and 10\% of the training data refer to subsets randomly 
sampled from the entire training dataset.}
\begin{tabular}{@{\hskip 20pt}c@{\hskip 20pt}l@{\hskip 20pt}c@{\hskip 20pt}@{\hskip 20pt}c@{\hskip 20pt}@{\hskip 20pt}c@{\hskip 20pt}}
\hline
Dataset & Method & 1\% & 5\% & 10\% \\
\hline
 \multirow{4}{*}{X-ray} 
 & Ours (uniform) & 0.750 $\pm$ 0.012 & 0.844 $\pm$ 0.004 & 0.871 $\pm$ 0.002\\
 & Ours (consecutive) & 0.745 $\pm$ 0.012 & 0.841 $\pm$ 0.004 & 0.869 $\pm$ 0.001 \\
 & Ours (random with consistency) & 0.746 $\pm$ 0.012 & 0.842 $\pm$ 0.004 & 0.867 $\pm$ 0.010 \\
 & Ours (random w/o consistency) & 0.742 $\pm$ 0.011 & 0.839 $\pm$ 0.004 & 0.864 $\pm$ 0.002 \\
\hline
 \multirow{4}{*}{CT} 
 & Ours (uniform) & 0.736 $\pm$ 0.007 & 0.828 $\pm$ 0.005 & 0.889 $\pm$ 0.003 \\
 & Ours (consecutive) & 0.729 $\pm$ 0.012 & 0.825 $\pm$ 0.005 & 0.884 $\pm$ 0.004 \\
 & Ours (random with consistency) & 0.730 $\pm$ 0.012 & 0.826 $\pm$ 0.004 & 0.887 $\pm$ 0.002 \\
 & Ours (random w/o consistency) & 0.727 $\pm$ 0.013 & 0.825 $\pm$ 0.007 & 0.886 $\pm$ 0.003 \\
\hline
 \multirow{4}{*}{MRI}
 & Ours (uniform) & 0.665 $\pm$ 0.030 & 0.774 $\pm$ 0.019 & 0.848 $\pm$ 0.016  \\
 & Ours (consecutive) & 0.664 $\pm$ 0.029 & 0.769 $\pm$ 0.015 & 0.846 $\pm$ 0.014 \\
 & Ours (random with consistency) & 0.667 $\pm$ 0.032 & 0.773 $\pm$ 0.016 & 0.847 $\pm$ 0.013 \\
 & Ours (random w/o consistency) & 0.665 $\pm$ 0.029 & 0.771 $\pm$ 0.028 & 0.843 $\pm$ 0.014 \\
\hline
\end{tabular}
\label{tab7}
\end{table*}
\begin{table}[t]
\centering
\small
\caption{GPU memory cost, training time per epoch, computational complexity (FLOPs), and inference speed (FPS) comparison. The training data used was a 1\% X-ray dataset.}
\begin{tabular}{cccccc}
\hline
  Method & Memory (M) & Time (s) & FLOPs (G) & FPS (fps) \\
\hline
 Ours & 2,434 & 3.24 & 1.078 & 201.44  \\
 CM1 & 1,932 & 2.69 & 1.078 & 167.10  \\ \hline
\end{tabular}
\label{tab8}
\end{table}
To evaluate the effectiveness and flexibility of our framework, we conducted experiments that combine our dual-model weight selection method with multiple SKD techniques. Using ViT-S as the teacher model and ViT-T as the student model, experiments were conducted across chest X-ray, lung CT, and brain MRI datasets. Table~\ref{tab5} summarizes the classification accuracy for each method.
We compared the following three SKD strategies: (1) basic SKD, (2) class similarity-based knowledge distillation (CS-KD)~\cite{Yun_2020_CVPR}, and (3) teacher-free knowledge distillation (TF-KD)~\cite{yuan2020revisiting}. CS-KD enhances traditional SKD by modifying the soft labels using a class similarity matrix, allowing the student model to learn from both the correct and closely related semantic classes. This is especially useful in medical image tasks where the class boundaries are subtle. In contrast, TF-KD uses the student model's outputs from previous epochs as pseudo teachers. By referencing past predictions, the model encourages consistency and smooth knowledge transfer across training. Both outputs were softened using temperature scaling, and their KL divergence was used as a loss term.
Overall, all methods combined with weight selection achieved consistently high classification accuracy, confirming the robustness of our approach. Our results demonstrate that the proposed method remains effective regardless of the specific SKD strategy applied. This suggests that the proposed framework is highly adaptable and can be adapted to various distillation strategies. The consistent improvements over CM1 also highlight that combining weight selection with SKD effectively enhances the learning ability of small models, especially in low-data medical imaging scenarios.
\subsection{Analysis of Weight Selection Methods}
\label{4.6}
To investigate the influence of different weight selection strategies, we conducted experiments using ViT-S as the teacher model and ViT-T as the student model across three medical imaging datasets: chest X-ray, lung CT scan, and brain MRI. Table~\ref{tab7} summarizes the classification accuracies achieved by each method. The strategies include uniform selection, consecutive selection, random selection with consistency, and random selection without consistency.
\par
Uniform selection samples slices evenly across all weight groups or channels, resulting in balanced coverage. This is particularly beneficial in architectures with grouped components, such as multihead attention in ViTs or grouped convolutions in ResNeXt. Consecutive selection picks adjacent slices, which in grouped architectures often results in selecting entire groups while skipping others, causing imbalance. Random selection with consistency applies the same randomly generated indices to all weight tensors, maintaining consistent positions across layers. Conversely,  random selection without consistency chooses indices independently for each tensor, leading to variations in selected positions. Among these methods, the proposed approach, along with uniform and consecutive selection, maintains index consistency across tensors. In contrast, random selection without consistency does not, which may negatively influence structural alignment.
\par
Experimental results reveal that although all methods perform similarly overall, the proposed method with uniform selection consistently achieves slightly higher accuracy. This indicates that maintaining consistent weight positions while sampling uniformly helps preserve the teacher model’s structural relationships, enabling more effective knowledge transfer and improved student model performance.
\begin{figure*}[t]
    \centering
    \includegraphics[width=18cm]{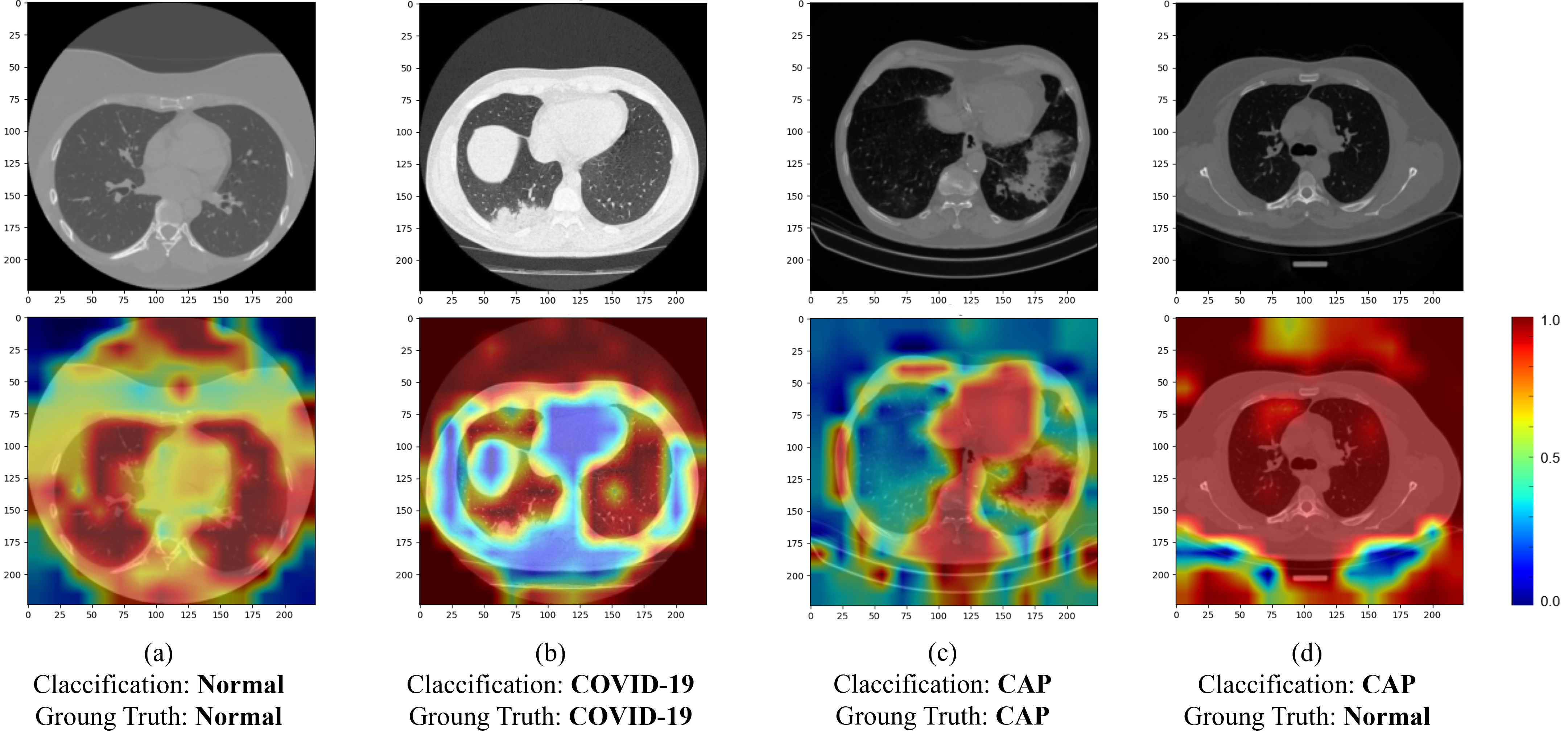}
    \caption{Grad-CAM visualization results on CT images. 
  From left to right, three correctly classified samples and one misclassified sample are shown. For each image, the upper label indicates the classification result, and the lower label indicates the ground truth. Red regions correspond to areas where the model exhibits close attention. In successful cases, the model accurately focused on lesion-related or anatomically relevant regions, whereas in the misclassified case, attention was diffusely distributed across irrelevant areas, indicating that the model failed to capture discriminative features.}
    \label{fig4}
\end{figure*}
\subsection{Computational Efficiency Analysis}
We compared the computational cost and training efficiency of the proposed method and CM1 on the chest X-ray dataset. To ensure a fair comparison, both methods employed ViT-S as the teacher model and ViT-T as the student model. Table~\ref{tab8} presents the GPU memory consumption, training time per epoch, theoretical computational complexity (FLOPs), and inference throughput (FPS).
\par
The proposed method introduces a modest increase in GPU memory usage and a slightly longer training time per epoch. However, this overhead is minor and remains well within a practical range. Also, the method consistently achieves higher classification accuracy on the chest X-ray dataset. In terms of inference efficiency, both methods share identical theoretical FLOPs, indicating that they require the same nominal computation. The slight reduction in FPS observed in our method is due to the auxiliary-model interaction introduced by the self-knowledge distillation process, which incurs minimal runtime overhead without altering the model architecture. Nonetheless, the inference speed remains comparable to CM1, demonstrating that the proposed method maintains real-world deployability while providing improved predictive performance.
\subsection{Visualization of Misclassified Samples}
To further examine model behavior beyond aggregate performance metrics, we analyzed several misclassified cases using Grad-CAM, as shown in Fig.~\ref{fig4}. CT images were selected for visualization due to their clearer structural contrast, which facilitates the interpretation of the highlighted regions. In successful classifications, the model consistently focused on lesion-related or anatomically relevant regions, reflecting meaningful feature localization. For example, in CAP cases, attention was concentrated in bright lesion regions, while for Normal and COVID-19 cases, the model emphasized darker lung areas. In contrast, misclassified samples exhibited diffuse or misplaced attention, indicating difficulty in identifying discriminative structural cues. These observations suggest that failure cases may arise from low-contrast regions or overlapping texture patterns, where the distinction between classes becomes less pronounced, and the model’s decision boundaries are less reliable. Notably, we did not observe consistent associations between these misclassifications and specific weight subset configurations, suggesting that the errors arise primarily from image-level ambiguity rather than the dual-model selection or distillation strategy itself.
\section{Discussion}
\label{sec5}
Our experimental results on three diverse medical imaging datasets---chest X-ray, lung CT scan, and brain MRI---demonstrate the effectiveness, robustness, and versatility of the proposed method, which combines dual-model weight selection with SKD. Compared to traditional weight initialization schemes and existing weight selection baselines, our approach consistently achieves superior classification accuracy, especially in low-data scenarios. This underscores its suitability for medical applications where labeled data are often scarce.
\par
Nonetheless, certain limitations remain. First, while our method demonstrates robustness across three imaging modalities, its performance under significant domain shifts such as unseen scanner types and varying acquisition protocols has not been fully explored. Future work will examine the performance of the proposed method under domain shift conditions, such as unseen scanner types and variations in acquisition protocols, to further assess its robustness in real-world clinical environments.
Next, in real-world clinical datasets, annotation noise arising from inter-observer variability, ambiguous findings, and report-derived labels is almost unavoidable and can degrade model performance. Although we do not explicitly simulate label noise in our experiments, two components of the proposed framework are expected to confer a certain degree of robustness: (i) the dual-model weight selection favors parameter configurations that generalize better on held-out data, implicitly reducing overfitting to mislabeled samples; and (ii) the SKD stage leverages soft self-predictions as complementary supervision, which can smooth out isolated noisy labels and encourage consistency across related classes. A more systematic study under controlled noise levels and realistic noise patterns remains an important direction for future work.
Also, we report the mean ± standard deviation over five independently sampled training subsets to assess performance stability and robustness. While these results show consistent improvements over the baseline, we acknowledge that formal statistical significance testing could further strengthen the conclusions and plan to address this in future work.
Additionally, we plan to extend our comparative analysis to include more advanced lightweight architectures, such as MobileNet, EfficientNet, and distilled transformer variants, to more clearly position the proposed method within the broader landscape of contemporary lightweight deep learning approaches.
Furthermore, we acknowledge that excessive correlation between the main and auxiliary student models may reduce the benefit of complementary representation learning, and exploring additional diversification strategies constitutes an important direction for future work.
\section{Conclusion}
\label{sec6}
In this work, we introduced a dual-model weight selection framework combined with self-knowledge distillation for medical image classification. By constructing complementary parameter subsets and employing an EMA-based auxiliary teacher, the proposed method enhances the representation capability of compact models while maintaining low computational cost. Experiments across three imaging modalities, i.e., chest X-ray, lung CT, and brain MRI, demonstrated consistent improvements in classification performance and stability over baseline approaches, highlighting the effectiveness and practicality of the framework in resource-constrained clinical scenarios. While the current evaluation focuses on publicly available datasets, further investigation is needed to assess the robustness of models under domain shifts, such as variations in scanner types and acquisition protocols. In future work, we aim to extend the method to additional medical imaging tasks (e.g., ultrasound and histopathology), explore adaptive strategies for dynamic subset construction, and perform cross-center validation to evaluate generalizability more comprehensively.

\section*{CRediT Authorship Contribution Statement}
Ayaka Tsutsumi: Writing–review \& editing, Writing–original draft, Visualization, Validation, Software, Methodology, Investigation, Formal analysis, Conceptualization. Guang Li: Writing–review \& editing, Supervision, Project administration, Methodology, Conceptualization, Formal analysis, Data curation, Funding acquisition. Ren Togo: Writing–review \& editing, Supervision, Conceptualization, Funding acquisition. Takahiro Ogawa: Writing–review \& editing, Supervision, Conceptualization, Funding acquisition. Satoshi Kondo: Writing–review \& editing, Supervision. Miki Haseyama: Supervision, Funding acquisition.

\section*{Ethics in Publishing Statement}
This study does not require an Ethics Statement, as it involves no new experiments on human or animal subjects. All analyses were performed using either simulated data based on publicly available sources or real-world expression datasets from previously published studies. These datasets were originally collected under appropriate ethical standards, with the necessary approvals obtained at the time. No personally identifiable or sensitive information was used in this research. Therefore, no additional ethical approval was needed for the present computational and inferential analyses.

\section*{Declaration of Competing Interest}
The authors declare no competing financial interests or personal relationships that could have influenced the work presented in this paper.
\section*{Acknowledgments}
This research was supported in part by JSPS KAKENHI Grant Numbers JP23K11141, JP23K21676, JP24K02942, JP24K23849, and JP25K21218.
\bibliographystyle{elsarticle-num}
\bibliography{CIBM}

\end{document}